%% file: domain-style-frame.tex
\documentclass[11pt,a4paper]{article}
\usepackage[american]{babel}
\usepackage[utf8]{inputenc}
\usepackage[T1]{fontenc}
\usepackage[hyperref]{acl2020}
\usepackage{times}
\usepackage{microtype}

\usepackage{diagbox}

\aclfinalcopy 

\usepackage{url}
\RequirePackage{type1cm}
\RequirePackage{soul}
\setstcolor{blue}

\usepackage{graphicx}
\DeclareGraphicsExtensions{.ai,.pdf,.jpg,.png}
\setkeys{Gin}{pagebox=artbox}
\graphicspath{{./figures/}}

\usepackage{subcaption}

\usepackage{booktabs}
\newcommand{\Ni}{(1)~}
\newcommand{\Nii}{(2)~}
\newcommand{\Niii}{(3)~}
\newcommand{\Niv}{(4)~}

\usepackage{xspace}

\newcommand{\tfidf}{\ensuremath{\mathit{tf}\kern-0.15em\cdot\kern-0.15em\mathit{idf}}}

\hypersetup{draft}

\begin{document}

\input{domain-style-pre}
\input{domain-style-part1}
\input{domain-style-part2}
\input{domain-style-part3}
\input{domain-style-part4}
\input{domain-style-part5}
\input{domain-style-part6}
\input{domain-style-sum}


\bibliographystyle{acl_natbib}
\bibliography{acl20-domain-invariant-style-representation-lit}


\end{document}

%% file: domain-style-pre.tex
\title{Domain-Adversarial Training for (Domain-Invariant) Writing Style Representations}
\title{Learning Domain-Invariant Writing Style Representations}
\title{Learning Writing Style Embeddings: Domain-Adversarial Training}
\title{Learning Writing Style Representations}
\title{Learning Writing Style Representations via Domain-Adversarial Training} 
\title{Heuristics vs. Domain-Adversarial Learning to Represent Writing Style}
\title{The Importance of Suppressing Domain Style in Authorship Analysis}

\author{
Sebastian Bischoff $^1$
\qquad Niklas Deckers $^2$
\qquad Marcel Schliebs $^3$
\qquad Ben Thies $^2$ \\[1ex]
{\bf
\qquad Matthias Hagen $^4$
\qquad Efstathios Stamatatos $^5$
\qquad Benno Stein $^6$
\qquad Martin Potthast $^7$} \\[2ex]
$^1$Technical University of Munich, {\small\tt sebastian.bischoff@tum.de} \\
$^2$Humboldt-Universit{\"a}t zu Berlin, {\small\tt \{niklas.deckers | ben.thies\}@hu-berlin.de} \\
$^3$University of Oxford, {\small\tt marcel.schliebs@oii.ox.ac.uk} \\
$^4$Martin-Luther-Universit{\"a}t Halle-Wittenberg, {\small\tt matthias.hagen@informatik.uni-halle.de} \\
$^5$University of the Aegean, {\small\tt stamatatos@aegean.gr} \\
$^6$Bauhaus-Universit{\"a}t Weimar, {\small\tt benno.stein@uni-weimar.de} \\
$^7$Leipzig University, {\small\tt martin.potthast@uni-leipzig.de}}

\date{}

\maketitle

\begin{abstract}
The prerequisite of many approaches to authorship analysis is a representation of writing style. But despite decades of research, it still remains unclear to what extent commonly used and widely accepted representations like character trigram frequencies actually represent an author's writing style, in contrast to more domain-specific style components or even topic. We address this shortcoming for the first time in a novel experimental setup of fixed authors but swapped domains between training and testing. With this setup, we reveal that approaches using character trigram features are highly susceptible to favor domain information when applied without attention to domains, suffering drops of up to~55.4 percentage points in classification accuracy under domain swapping. We further propose a new remedy based on domain-adversarial learning and compare it to ones from the literature based on heuristic rules. Both can work well, reducing accuracy losses under domain swapping to~3.6\% and~3.9\%, respectively.
\end{abstract}

%% file: domain-style-part1.tex
\section{Introduction}

Authorship analysis refers to a group of tasks in which authors are modeled based on their writing style. The most commonly studied task is authorship attribution, where for a text of unknown authorship it has to be decided who of a given set of candidate authors has written it. Under the assumption that every author possesses a unique writing style, unconsciously encoded into their writing, the attribution is solved by picking the candidate whose writing style best matches the one found in the text in question. Unlike a text's topic, however, writing style has proven difficult to be modeled reliably---the key challenge of any authorship analysis.

The style of a text as perceived by human readers results not only from an author's personal traits, but also from customs an author adopts due to genre, register, type, and topic. These concepts are vague, and each can be defined broadly but also subdivided hierarchically, rendering them difficult to be operationalized. Style forms a continuum, and the challenge is thus to discover (combinations of) style markers more likely to be determined by an author's personality rather than by domain customs.

Due to the lack of large-scale datasets, most machine learning approaches to authorship attribution are still based on manual feature engineering. It is commonly agreed to avoid certain features like content words, which rather capture topic than style. Yet, style-capturing features are mostly based on intuition about what (combination of) quantifiable characteristic(s) might represent an author. While many have argued why their features should capture author style, such claims are hardly ever substantiated experimentally: Typically, experiments do not control for domain-related style components, foreclosing conclusions about the true capabilities of a feature set in capturing author style. What is worse, a feature set that captures author style more ``clearly'' may even go unremarked compared to one that also captures other domain characteristics, since, in the typical experimental setups, the latter has a better chance of performing well.

Our contributions address these shortcomings for the first time:
\Ni
We devise the first evaluation setup to explicitly measure the capabilities of style representations (Section~\ref{measuring-author-style}).
\Nii
To enable many corresponding experiments as well as the application of deep learning, we compile a large corpus of 1.4~million authors, each of whom has written long monographs (5.8~million stories in 44~languages) in 10,328~different topical domains (Section~\ref{corpus-construction}).
\Niii
We apply domain-adversarial training to train the first neural style encoder that suppresses domain-specific information (Section~\ref{style-encoder}).
\Niv
In a series of experiments, the new style encoders are compared to competitive baselines, showing that traditional character trigram models are extremely susceptible to capturing domain style instead of author style, whereas domain-adversarial learning and heuristic rules are not (Section~\ref{evaluation}).

%% file: domain-style-part2.tex
\section{Related Work}

Given the large number of papers about authorship analysis in general and attribution in particular (extensively surveyed by \citet{stamatatos:2009} and \citet{neal:2018}), we focus on cross-domain attribution and recent related deep learning advances.

\subsection{Cross-Domain Attribution}

Forensic linguists hardly ever encounter attribution problems where the domains of the text of unknown authorship and the writing samples of the candidate authors are the same. Despite its practical importance, however, cross-domain attribution has been studied much less compared to same-domain attribution.
\citet{mikros:2007} has been the first to investigate the topic dependence of features on a dataset of 200~Greek newspaper articles from two topics. Features like frequent function words and word length had a high correlation with topic; character trigrams were not studied. \citet{stamatatos:2013} later showed character trigrams to be highly effective on 1004~articles from The Guardian across two genres and four topics. In a follow-up, \citet{stamatatos:2017} suggested to reduce domain-specific information through a text distortion phase before the feature extraction; one of our baselines.

\citet{kestemont:2018} first proposed fanfiction as a source for cross-domain authorship problems. For their shared task at~PAN, a sample of 40~attribution problems was compiled ranging from~5 to 20~candidate authors with 7~texts each. The best-performing approach relied on an ensemble classifier based on character n-grams and the text distortion approach of \citet{stamatatos:2017}. We scale this idea by crawling a fanfiction corpus comprising more than a million authors who wrote millions of stories across thousands of domains.

Besides topic, also genre and language have been investigated as domain variables. \citet{stamatatos:2013} reports character trigrams to perform well across two genres---yet worse than for cross-topic attribution---and \citet{overdorf:2016} report a drop of attribution performance across three social media genres compared to same-genre attribution.
\citet{bogdanova:2014} study attribution across languages, combining machine translation with cross-language features,
and \citet{vandergootBleachingTextAbstract2018} improve cross-language gender-prediction by bleaching text through transforming lexical strings into more abstract
features.

Altogether, none of the above studies shed light on the question how well author style is separated from domain style components.

\subsection{Deep Learning-Based Attribution}

A handful of attribution studies investigated the usefulness of deep learning. \citet{ruder:2016} outperform several traditional state-of-the-art attribution approaches in same-domain settings using convolutional neural networks~(CNNs). However, they caution that ``fine-tuned word embeddings that are sensitive to topical divergence between authors boost CNN~performance.'' \citet{boumber:2018} propose another CNN~approach designed for multi-label attribution tasks, but also take advantage of topic information through word embeddings.

\citet{hassan:2017} achieve 95\%~attribution accuracy on scientific papers via a supervised~LSTM and lexical and syntactic features. However, since topic seems to not have been controlled during training, it is unclear whether writing style was actually learned. Recently, \citet{ding:2019} suggested a representation learning approach using a neural network that combines character, lexical, syntactical, and topical information. Again, an analysis of how well the representations separate an author's writing style from domain style was beyond the scope of their study, which seems problematic since topic information was exploited.


\subsection{Style Transfer}

Given the success of style transfer for images (e.g., \citealp{gatys:2016}), many studies also try to translate or rather paraphrase a text from a source style to a desired target style, the two main problems being the lack of large-scale parallel training data (i.e., texts written in both styles), and the lack of reliable evaluation metrics (i.e., humans need to assess the transfer quality) \citep{fu:2017}. Style transfer studies typically do not model an author's personal writing style, but consider more broad style characteristics like a text's sentiment \cite{hu:2017,shen:2017,zhang:2018}, dialects of English (formal vs.\ informal, Shakespearean vs.\ simple) \cite{jhamtani:2017,jin:2019,kabbara:2016}, or political slant \cite{prabhumoye:2018}. Some text style transfer studies even suggest not to disentangle latent representations of style and content \cite{dai:2019}---exactly the opposite of what we require from a writing style representation for attribution.

\subsection{Adversarial Training}

In author obfuscation, the task is to paraphrase a given text to render an author's style imperceptible~\cite{stein:2016k}. Typically, another text from the author is used as a reference for style similarity; recent approaches employ neural models \cite{emmery:2018} and heuristic search \cite{stein:2019l} to render the given text dissimilar. Similarly, \citet{elazar2018adversarialremovaldemographic} attempt to remove markers from a style representation to protect its author's demographic details, such as gender, age, etc., through adversarial training. Furthermore, \citet{griesshaberLowresourceTextClassification2020} use adversarial learning as a regularizer to avoid overfitting when training features for deep neural networks in low-resource settings.

In this paper, we also tackle cross-domain attribution with adversarial learning. We repurpose the adversarial transfer learning approach by \citet{ganin:2016}, which, by training on labeled data and adversarial training on unlabeled data, promotes features that are discriminative for the intended learning task but at the same time indiscriminative for the differences between the labeled and the unlabeled data, thus enabling a robust transfer. We observe that cross-domain attribution may be tackled in a similar fashion: by training on the texts with respect to their author labels, and adversarial training on the texts with respect to their domain labels, our approach promotes features that are discriminative for the task of authorship attribution but at the same time indiscriminative for the text domain differences. We adapt and improve the architecture to obtain substantial improvements, yielding effective cross-domain writing style representations.

%% file: domain-style-part3.tex
\section{Measuring Author Style}
\label{measuring-author-style}

How can the capabilities of a writing style model in capturing author style be reliably measured? The most commonly carried out experiment in the literature answers this question only under near-perfect conditions, but may otherwise yield misleading results. We argue that a careful control of the text domain is less error-prone and more insightful.

\subsection{Constructing Attribution Problems}

An authorship attribution problem consists of a text~$x$ of unknown authorship, and $k>1$ texts from known candidate authors, where~$x$ is to be attributed to the candidate whose writing style it matches. A typical scheme for problem instances for experiments for~$k=2$ authors looks as follows:

\smallskip
\begin{center}
\small
\setlength\tabcolsep{5pt}
\begin{tabular}{@{}r@{\qquad}cc@{\qquad}cc@{}}
\toprule
Scheme $S_1$ & \multicolumn{2}{@{}c@{\qquad}}{training} & \multicolumn{2}{@{}c@{}}{testing} \\ \midrule
authors & A      & B      & A       & B      \\
domains & P      & Q      & P       & Q      \\
\bottomrule
\end{tabular}
\end{center}

\smallskip\noindent
where~A,~B are authors and~P,~Q domains, and the vertical mapping denotes which author has written in which domain. For training, texts from~A and~B take turns as~$x$; for testing, previously unseen texts from~A and~B are used as~$x$. This scheme readily extends to $k>2$ authors.

The vast majority of experiments in the literature are within-domain, i.e., P~$=$~Q. Here, ensuring that all texts are mutually from the same domain includes checking their topic, genre, register, idiolect, time period, etc. \cite{grieve:2007}. The rationale is to ensure that style characteristics whose variation is due to domain style rather than author style are randomly distributed across the texts of all authors. Otherwise, latent domain differences may bias a model trained on a given style representation. In that case, the catch is that the biased model performs {\itshape better} on the test data, not worse, since it exploits domain style on top of author style.

The efforts that must be taken to properly construct a within-domain attribution problem should not be underestimated. Take genre as an example domain, which is only vaguely defined and hierarchical in nature: restricting texts to fiction may not be enough if~A is a romance author and~B a thriller author. Outside fiction, such problems take different forms: for instance, person~A may write mostly work-related emails, and~B personal ones. Ensuring domain equality with respect to genre as well as the other domains presumes in-depth knowledge of each individual text, severely limiting scalability.

\subsection{Domain Swapping vs. Author Style}
\label{sec:swapping}

To explicitly quantify the capabilities of a style model in capturing author style, we propose to contrast the performance achieved with Scheme~$S_1$ with that of the following:

\smallskip
\begin{center}
\small
\setlength\tabcolsep{5pt}
\begin{tabular}{@{}r@{\qquad}cc@{\qquad}cc@{}}
\toprule
Scheme $S_2$ & \multicolumn{2}{@{}c@{\qquad}}{training} & \multicolumn{2}{@{}c@{}}{testing} \\ \midrule
authors & A      & B      & A       & B      \\
domains & P      & Q      & Q       & P      \\
\bottomrule
\end{tabular}
\end{center}

\smallskip\noindent
where P~$\neq$~Q and the relation between authors and domains is swapped between training and test, which we call {\itshape domain swapping}. Given a style model and a performance measure, by computing the difference $\Delta(S_1,S_2)$ of the performance the model achieves in experiments as per Schemes~$S_1$ and~$S_2$, one can directly observe the proportion of performance a model achieves due to exploiting domain style as opposed to author style.
In this cross-domain experiment, the requirement of within-domain experiments to tightly control all conceivable domains that may affect style besides the author is relaxed: A style model and the data used to train it may be developed domain by domain, allowing for incremental improvements.

%% file: domain-style-part4.tex
\section{Attribution Problems from Fanfiction}
\label{corpus-construction}

Following \citet{kestemont:2018}, we employ fanfiction---fiction written by fans of another's work, reusing its characters and settings---as a large-scale source of ground truth. The original work a given fanfiction is about is called its fandom. For well-known fandoms, such as Harry Potter, many different authors have written fanfiction. Moreover, many authors contribute to more than one fandom. We operationalize fandoms as (topic) domains in our experiments. Each fandom refers to a different ``universe'' of storytelling, quite distinct from others, yet also quite consistent within itself.

Unfortunately, the datasets provided by \citet{kestemont:2018} lack the required domain labels and comprise only a few hundred texts, so that we resorted to crawling \href{www.fanfiction.net}{fanfiction.net} instead. We rigorously cleaned the texts to remove any mentions of author names, notes, and disclaimers. Moreover, we excluded ``crossovers'' combining aspects from two or more fandoms. Altogether, we compiled 5,800,292 stories in 44~languages from 1,400,958 authors writing in a total of 10,328~different fandoms.%
This corpus is made available to also enable future cross-domain attribution research.

\subsection{Sampling Training and Test Data}

From this corpus we construct instances of attribution problems as outlined above. The amount of available problem instances depends on the desired experiment scheme as well as the following constraints: our focus is on the English subset of the corpus, and the most basic problems with two authors and two fandoms each. We leave experiments with more languages and authors to future work.

The input size of our neural style encoder is presently 500~words, which is frequently considered to be about the minimum sufficient length to measure author style (e.g.,~\citealp{koppel:2004}). Although it is possible in principle to train a style encoder with such a small amount of text per author per domain, in our pilot experiments, we determined 100,000 words per author per domain (in the form of 500~word chunks) to be a sensible lower bound (Section~\ref{traditional-attribution}). Moreover, to maximize reliable testing, we desire at least another 50,000 words per author per domain.
Additionally, we employ a validation set of 50,000 words per author per domain for hyperparameter optimization and early stopping on our neural networks. We also make sure to never split chunks from a single fanfiction between training, validation and testing set.
Regarding all of the above constraints for Scheme~$S_1$ with P~$\neq$~Q, the corpus allows for drawing (with replacement) a total of 1,260,082,646 problem instances. Regarding Scheme~$S_2$ with P~$\neq$~Q, a total of 93,238 problem instances can be drawn.
In our evaluation, each experiment is repeated with at least ten distinct pairs of P,~Q.

%% file: domain-style-part5.tex
\newcommand{\text}{x}
\newcommand{\texts}{X}
\newcommand{\textrep}{\mathbf{x}}
\newcommand{\textreps}{\mathbf{X}}
\newcommand{\authorlabel}{y_{a}}
\newcommand{\authorlabels}{Y_{a}}
\newcommand{\domainlabel}{y_{d}}
\newcommand{\domainlabels}{Y_{d}}

\newcommand{\stylerep}{w(\textrep)}
\newcommand{\stylereps}{\mathbf{W}}

\section{Domain-Invariant Style Encoder}
\label{style-encoder}

\input{figure-neural-network-architecture}

Let~$\text$ denote a text, $\textrep$~its two-dimensional representation, and~$\texts$ as well as~$\textreps$ the sets of texts and text representations, respectively. An author mapping $a:X\to\authorlabels$ maps texts to their authors, and a domain mapping $d:X\to\domainlabels$ to their domains.

In what follows, we detail the input text representation, the adversarial training architecture, two style encoder variants, how we deal with the unbalanced class distributions as well as how we combine author and domain loss functions.

\subsection{Input Text Representation}
\label{input-text-representation}

An input text~$\text$ is a token sequence of 500~tokens. It is represented as a 2013$\times$500 matrix~$\textrep$. The matrix is composed of 500~one-hot vectors, one for each token in~$\text$ with 2013~dimensions each. 2001~dimensions represent the most frequently used words in the Brown corpus~\cite{francis:1965}, 11~represent punctuation marks, and one one-hot vector $(1,0,0,\ldots)$ represents all other tokens in~$\text$. Beforehand, all punctuation marks are reduced to their ASCII equivalents.
We refrained from introducing variables for these figures: it is our current trade-off between the maximal text length~$\text$, the size of its representation~$\textrep$, and the batch size $m=400$ of text representations~$\textreps_\mathrm{batch}\subset\textreps$ that we could simultaneously fit into our graphics card's RAM for training. We sought to maximize the number of examples in a batch while allowing for a reasonably-sized representation of each individual text.

\subsection{Adversarial Learning Architecture}
\label{sec:adv_approach}

As illustrated in Figure~\ref{figure-neural-network-architecture}, our goal is to find a writing style encoder $w:\textreps\to\stylereps$ which maps text representations to style representations~$\stylereps$, so that an author classifier~$c_a:\stylereps\to\authorlabels$ is successful in mapping the style representations to authors, and a domain classifier~$c_d:\stylereps\to\domainlabels$ is {\em unsuccessful} in mapping the same style representations to domains.

We train our neural network batch-wise using $\textreps_\mathrm{train}\subset\textreps$ to predict both~$\authorlabels$ and~$\domainlabels$, minimizing the two classifier's loss functions:
\begin{eqnarray*}
\mathcal{L}_a(c_a(w(\textreps_\mathrm{train}, \theta_{w}), \theta_{a}), \authorlabels);\\[1ex]
\mathcal{L}_d(c_d(w(\textreps_\mathrm{train}, \theta_{w}), \theta_{d}), \domainlabels),
\end{eqnarray*}
where for $w$, $c_a$, and $c_d$ their respective parameters $\theta_w,\theta_a,\theta_d$ are updated with respect to their joint loss $w(\textreps_\mathrm{train},\theta_w)$, $c_a(\textreps_\mathrm{train},\theta_a)$, and $c_d(\textreps_\mathrm{train},\theta_d)$ for the training texts~$\textreps_\mathrm{train}$.

By setting up the backpropagation of the prediction losses for both classifiers so that the obtained prediction performance for the author labels~$\authorlabels$ is maximized while that for the domain labels~$\domainlabels$ is minimized, the encoder learns to encode author style while avoiding domain style. This is accomplished by reversing (negating) the gradient of the domain classifier when propagated back to the style encoder. A gradient descent of negated gradients equals a gradient ascent of the original gradients. Therefore, while the optimizer updates the weights in the domain classifier to better predict the domain, at the same time, the weights in the style encoder are updated such that the style vector becomes less helpful for domain prediction.

Once the neural network has been trained using~$\textreps$, we retain only its encoder: it takes a representation $\textrep$ of a given (previously unseen) text~$\text$ as input and derives the style vector $\stylerep$, which is well-suited to predict~$\text$'s author label~$\authorlabel$ but unsuited to predict~$\text$'s domain label~$\domainlabel$. The encoder suppresses domain style while retaining author style, rendering $\stylerep$ a domain-invariant author style vector for text~$x$.

Dropout~\cite{srivastava:2014} is used for regularization and its resemblance of ensemble learning~\cite{baldi:2013}. 
We use dropout on the input and all fully-connected layers.
Batch normalization~\cite{ioffe:2015} ensures better learning dynamics. This is especially important in our case because the interaction of the two different losses makes the problem hard to optimize. Contrary to the results of \citet{ioffe:2015}, we cannot reduce the dropout rate and still get the same performance. We use Adam~\cite{kingma:2014} as an optimizer with Xavier initialization~\cite{glorot:2010} for the output layers with sigmoid activation functions and He initialization~\cite{he:2015} for fully connected layers with ReLU activation functions.

\subsection{Style Encoder Variants}

We consider two variants of our style encoder. The first one (Encoder~1), also illustrated in Figure~\ref{figure-neural-network-architecture}, exploits the sequential nature of our text representation~$\textrep$ by feeding it into a recurrent neural network~(RNN) in the form of an~LSTM \cite{hochreiter:1997}. The representation resulting from the LSTM is then fed into a convolutional layer to obtain the writing style representation. Although this architecture does improve over a convolutional layer in isolation, we observe that only the loss of the representation originating from the LSTM's final step is taken into account.

As a second variant (Encoder~2), in order to extract more information for the optimization of the style encoder's weights, we do not only use the loss when predicting the author on the full text representation~$\textrep=\textrep_{1:500}$ (i.e., one-hot encoded token~1 to token~500), but the losses from~$n$ predictions based on the style representations obtained after the LSTM has read $\textrep_{1:\frac{1}{n}500}, \textrep_{1:\frac{2}{n}500}, \ldots, \textrep_{1:500}$.
The individual losses are combined as follows:
$$
\sum_{i=1}^n \lambda_i \mathcal{L}_a(a(w(\textreps_{1:\frac{i}{n}500}, \cdot), \cdot), \authorlabels),
$$
weighting the prediction on $\textrep_{1:\frac{1}{n}500}$ the least and the one on $\textrep_{1:500}=\textrep$ the most. This way of training our style encoder can also be viewed as parameter sharing between individual networks trained on texts of length $\frac{1}{n}500, \ldots, 500$.

\subsection{Unbalanced Author and Domain Classes}
\label{unbalanced}

Unbalanced class distributions are often balanced using approaches like oversampling an underrepresented class or undersampling an overrepresented class. The rationale is to balance the influence of each class (i.e., in terms of macro accuracy). However, these sampling techniques are not directly applicable to problems formed of pairs~$(\domainlabel, \authorlabel) \in (\domainlabels, \authorlabels)$ of labels. Changing the distribution of~$\domainlabels$ also changes that of~$\authorlabels$, and vice versa. We therefore use the macro author accuracy and correct the author loss by weighting the texts from author~A with ${m}/{n_\mathrm{A}}$, where $m$ is the number of authors and $n_\mathrm{A}$ is the number of texts written by author~A. Likewise, the domain accuracy and domain loss are calculated.

It should be noted that not only the total numbers of activity $n_\mathrm{A}$ and $n_\mathrm{D}$ vary, but also the activity $n_\mathrm{A,D}$ of authors~A in domains~D. While this does not affect the validity of the accuracy calculations, it may affect the effectivity of the adversarial learning: if the correlation between authors and domain is~1, i.e., if author labels can be mapped to domain labels, the performance of the author classifier also correlates with the performance of the domain classifier. This problem is outlined in the next section.

\subsection{Combining Author and Domain Loss}

To obtain a joint loss function $\mathcal{L}(\dots, (\authorlabels, \domainlabels))$, it might seem natural to add author loss and domain loss, since both losses are to be minimized. However, in our case, author and domain are correlated, which requires an adaptation of the loss function: Ideally, the domain accuracy is not reduced to~0, but to a random guess based on the writing style. This yields a lower bound of accuracy $1/n_\mathrm{D}$, which may be higher due to correlations between author and domain, presuming we achieve a high author accuracy. When simply adding the losses, the domain loss $\mathcal{L}_d(\ldots, \domainlabels)$ will be decreased below that boundary at the expense of author accuracy.

Instead, we employ another loss combination method based on the lower bound of the domain accuracy given the current author accuracy. When passing a single text through the current state of the network, it provides a vector $\mathbf{p}$ of prediction probabilities for each author. Assuming that all domain information is eliminated from the style representation, the author predictions might still be used to perform a prediction of the domains. The column-normalized matrix $\mathbf{N}$ of activity values $n_\mathrm{A,D}$ contains the conditional probabilities $P(D | A)$, i.e., the probability for each author~$A$ to write in some domain~$D$. This matrix is applied on the predicted author probabilities $P(A)$ to receive the unconditional domain probabilities $P(D)$ as a vector $\mathbf{q}$:
$$
P(D)=\sum_{A}{P(D | A)\cdot P(A)},\quad
\mbox{or}\quad
\mathbf{q}=\mathbf{N} \cdot \mathbf{p}.
$$
As the domain classifier is optimized, the network reaches a domain macro accuracy performing at least as good as the macro accuracy of the predictions implied by~$\mathbf{q}$ (calculated by selecting the most probable domain for each text). We call this lower bound~$b$. In order to perform as good as~$b$, the domain classifier reproduces the author classification and empirically estimates the matrix~$\mathbf{N}$.

\begin{figure}
\includegraphics[scale=0.45]{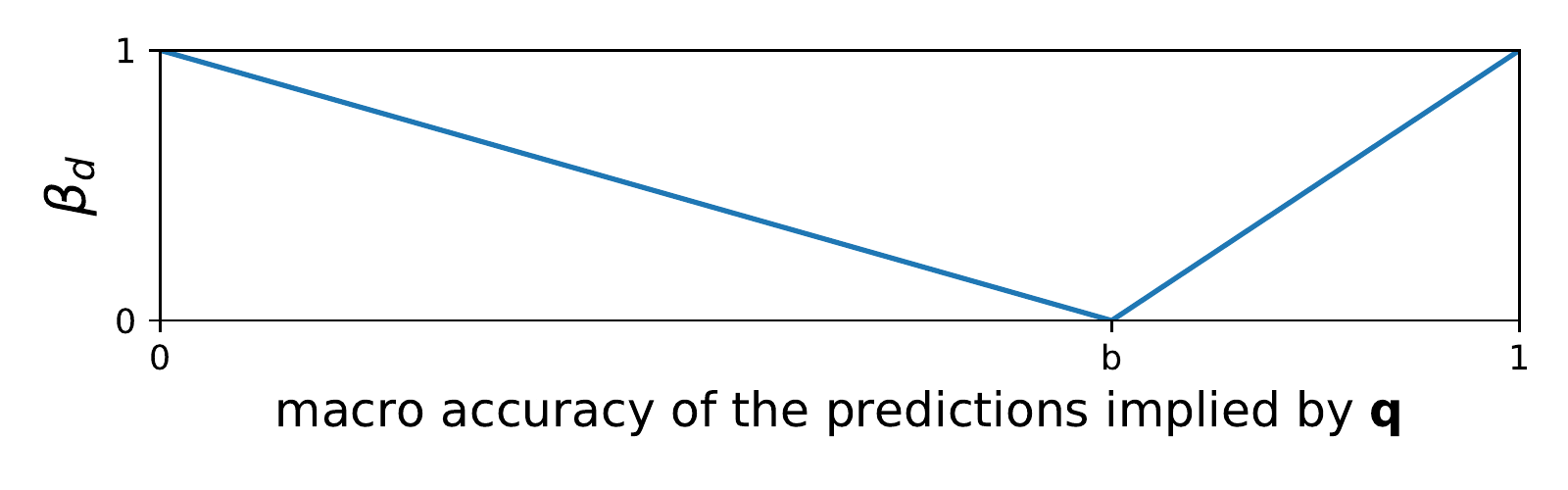}
\caption{Determination of $\beta_d$ for loss combination.}
\label{fig-loss}
\end{figure}

In mid-training of the network, the style representation might still contain some domain information. Thus, the domain macro accuracy of the network will be higher than~$b$. The larger this difference is, the more important it gets to eliminate domain information. This can be achieved by increasing the portion~$\beta_d$ of the domain loss according to Figure~\ref{fig-loss}. Finally, the combined loss is defined as an affine combination of author and domain loss:
\begin{eqnarray*}
\mathcal{L}(\dots, (\authorlabels, \domainlabels)) &=& \beta_d\ \mathcal{L}_d(\dots, \domainlabels) \\ 
&& +\ (1-\beta_d)\ \mathcal{L}_a(\dots, \authorlabels).
\end{eqnarray*}

%% file: figure-neural-network-architecture.tex
\begin{figure*}[t]%
\centering%
\includegraphics{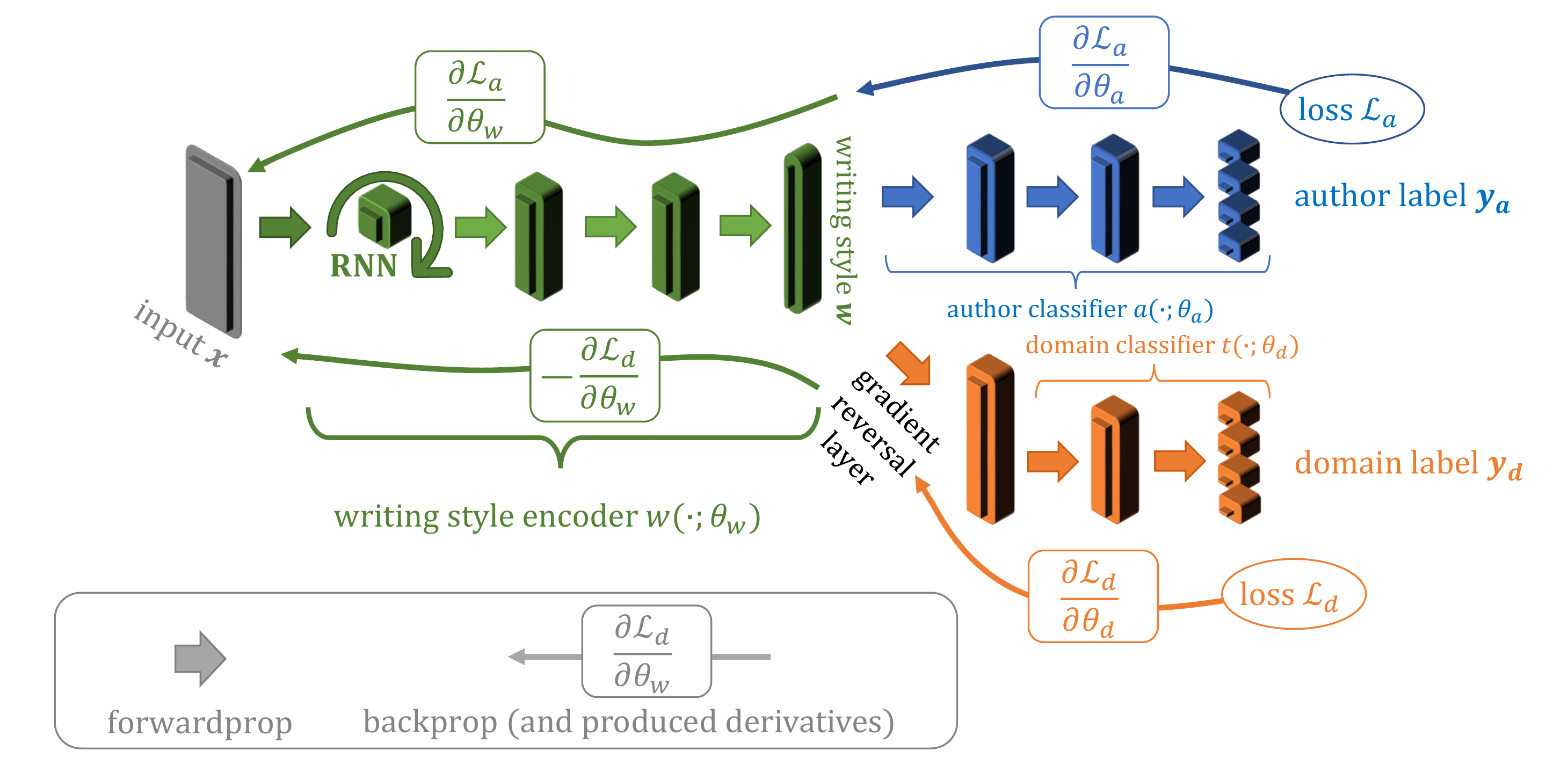}%
\caption{Architecture of our network including writing style encoder, author classifier and domain classifier. The texts are fed into the network as $\textrep$, a recurrent neural network is used to process the sequential property of a text. The final hidden state is used as a fixed-size summarization of the sequence~\cite[p.~371]{goodfellow:2016} for the following fully connected layers to yield the writing style representation~$\stylerep$ (with parameters $\theta_w$) of the full writing style encoder. The gradient can therefore only be obtained on the final hidden state which makes it difficult to optimize. This text representation is fed into the author classifier~$c_a$ (parameters~$\theta_a$) and domain classifier~$c_d$ (parameters~$\theta_d$). When backpropagating, we reverse the gradients for~$\theta_d$ before the writing style encoder.}%
\label{figure-neural-network-architecture}%
\end{figure*}

%% file: domain-style-part6.tex
\input{table-evaluation-results}

\section{Evaluation}
\label{evaluation}

This section reports on a series of experiments to study whether and to what extent character trigram representations capture domain style information, and, whether and to what extent domain style can be successfully suppressed. Regarding the latter, we compare heuristic rules that have been applied in the literature with our new domain-adversarial learning approach and our two writing style encoder variants. With our experiment series, we want to shed light on the following questions:
\begin{enumerate}
\item
Does domain-adversarial learning compete with traditional approaches to authorship attribution in a traditional setting?
\item
What is the impact of domain swapping on all models under investigation? In particular: Can domain style be successfully suppressed?
\item
What is the generalizability of our style encoders across fandoms?
\end{enumerate}

\paragraph{Experimental Setup}

The experiments pertaining to Question~1 establish that all models under investigation are on par with each other. This includes the reconciliation of the training requirements of the different machine learning paradigms. Key goal is the creation of a setup in which all models are provided with the same amount of information for training. Regarding the experiments pertaining to Question~2, achieving this kind of fairness while at the same time meeting the requirements of the domain swapping experiment {\em and} those of domain-adversarial learning in terms of scale of training data presented a unique challenge.  With all constraints combined, the fanfiction data we compiled provided for a sufficient amount of training and test data, allowing for ten repetitions of every experiment. 

The models under investigation include:
\Ni
A standard character trigram representation in conjunction with the three learning algorithms support vector machine~(SVM), naive Bayes~(NB), and random forest~(RF).
\Nii
A reproduction of the rule-based domain style suppression approach by \citet{stamatatos2018masking}, which uses character trigrams after applying text distortion algorithms: replacing tokens with multiple asterisks~(MA) as per their length, or with a single asterisk~(SA); retaining only exterior characters~(EX) of words in a dictionary, or their last two~(L2) characters.
\Niii
The two writing style encoder variants (Encoder~1 and Encoder~2) introduced above.
In pilot experiments, all models haven been meticulously optimized with regard to their respective parameters.

As performance measure, we employ the mean macro accuracy over at least ten problem instances for every experiment.

\begin{figure}
\centering
\includegraphics[scale=0.4]{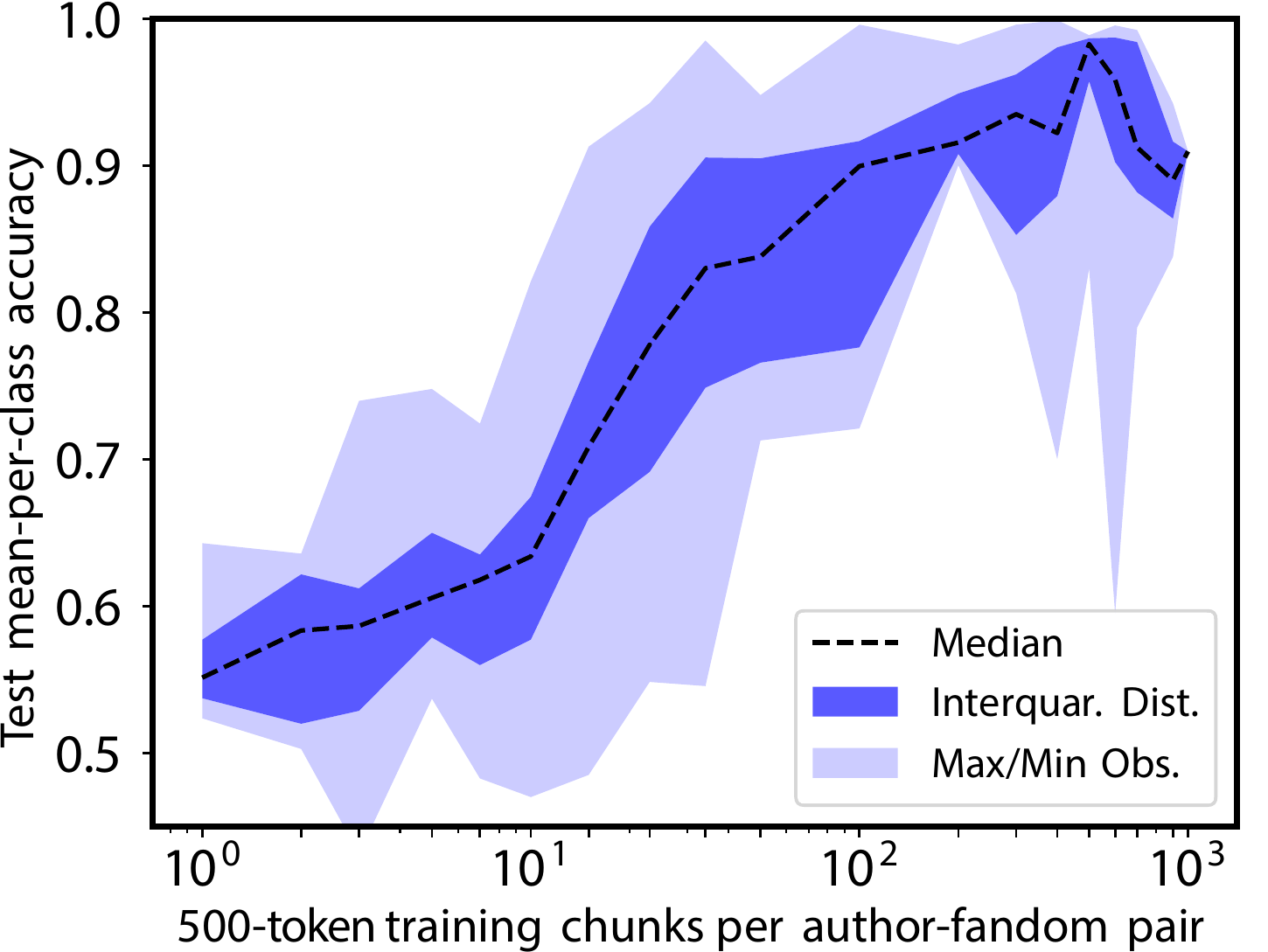} 
\caption{Author macro accuracy for test set over several text lengths.}
\label{fig:word-acc}
\end{figure}

\subsection{Traditional Authorship Attribution}
\label{traditional-attribution}

This experiment investigates the performance of all models within the following setup:

\input{table-experiment1-schema}

In this setup, two authors~A and~B have written in two fandoms~P and~Q, and an equal amount of their writing in both fandoms is used for training and testing, which corresponds most closely to a traditional attribution experiment from the literature. However, to allow for domain-adversarial learning, two fandoms must be present. Recall that this setup has been instantiated ten times without replacement from our fanfiction corpus for different pairs of authors~A,~B, and fandoms~P,~Q.

Important variables in this regard are the size~$|x|$ of an individual text~$x$ used for training or test, and the number~$|X_\mathrm{train}|$ of such texts~$X_\mathrm{train}$ per author and per fandom. Regarding the former, due to the constraints imposed by the adversarial learning approach (see Section~\ref{input-text-representation}), $|x|$~cannot exceed 500~tokens, which must be propagated to all other models for reasons of fairness. Regarding the latter, however, deep learning models require large amounts of training data when trained from scratch, which, too, must also be extended to all other models. To determine the amount of text required to reliably train our model, we train and test it on varying numbers of text chunks of 500~token each. Using the above experimental setup, we determine the macro accuracies over the number $|X_\mathrm{train}|$ of available chunks for training. The results can be seen in Figure~\ref{fig:word-acc}, where mean performance exceeds 90\%~accuracy at 200~chunks (100,000~words, the length of a book), which we choose as least amount of training text for each problem instance drawn from our corpus in all subsequent experiments, equally distributed across authors and fandoms.

\begin{figure}
\centering
\includegraphics[scale=0.5]{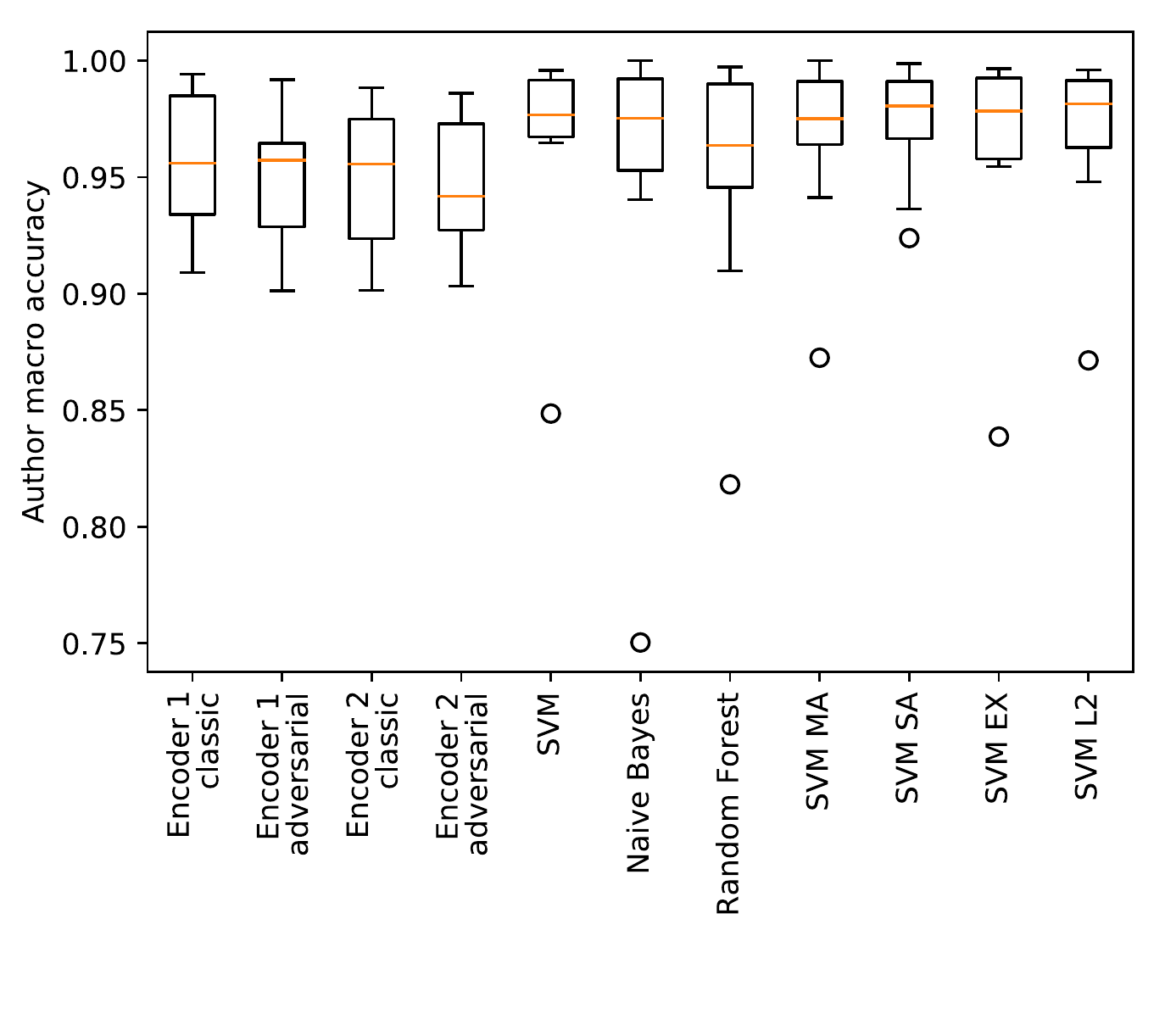}
\caption{Distribution of accuracies as per experiment Section~\ref{traditional-attribution}.}
\label{fig:whisker-attribution_using_adversarial_approach}
\end{figure}

Table~\ref{table-evaluation-results}a shows the accuracies of all models in this experimental setup, and Figure~\ref{fig:whisker-attribution_using_adversarial_approach} the respective performance distributions. We distinguish models that actively suppress domain style from ones that do not~(default). For Encoders~1 and~2, we disabled adversarial learning once to also supply performance values in a default learning setup. Otherwise, we observe that our fandom prediction accuracy converges to~50\% in adversarial learning in about two hours on a single GTX~1080.%

As can be seen, all models achieve very good accuracies between~94\% and~97\%. This performance is due to the amount of training data available; furthermore, the commonly held belief that 500~token chunks are sufficient to measure style is corroborated. Interestingly, neither does adding heuristic rules for domain suppression to the SVM change its a priori performance a lot (compare SVM with its variants MA, SA, EX, and L2), nor does adding adversarial training affect the performance of our writing style encoder. Despite their slight advantage in mean performance, the SVM models have a higher chance of severe misclassification, as the outliers in Figure~\ref{fig:whisker-attribution_using_adversarial_approach} show. As outlined in Section~\ref{measuring-author-style}, a setup like this does not reveal whether the performance difference is due to more or less usage of domain style information as opposed to author style, so that no conclusion about the relative effectiveness of the two suppression paradigms (heuristic rules vs.\ adversarial learning) can be drawn. Likewise, one cannot conclude that SVM-based models work ``better'' than our style encoder in terms of representing author style, since we cannot rule out that the better-performing models only perform better due to exploitation more domain style information.

\subsection{Domain Swapping}
\label{sec:exp2}

This experiment investigates the performance of all models within the following setup:
\input{table-experiment2-schema}

It contrasts a traditional authorship attribution situation with our novel domain swapping experiment. Two authors~C and~D have written texts in both domains~R and~S. For training, the models can learn from only one relation between authors and domains C-R and D-S, whereas for testing, either the same relation is used (``normal''), or a swapped relation C-S and D-R. This way, model bias that is due to an undesired exploitation of domain style rather than the desired representation of author style can be measured.

\input{table-correlation-changes}

We consider two kinds of domain swapping experiments:
\Ni
zero-knowledge swapping, and
\Ni
high-imbalance swapping.
The first variant, as shown in Table~\ref{table-correlation-changes}, maximizes the potential for confusion during training: the models never see an author in writing in the other author's respective fandom. However, this setup forecloses domain-adversarial learning, since the adversarial component cannot be trained in the absence of information about the domain to be suppressed. The second variant, as shown in Table~\ref{table-correlation-changes-relaxed}, relaxes the first variant by allowing for many examples of one author-fandom relation and only a few ones of the reverse relation during training, while swapping the imbalance for testing. This allows for adversarial training while approximating zero-knowledge swapping.

\input{table-correlation-changes-relaxed}

\begin{figure}[t]
\begin{subfigure}{.5\textwidth}
\centering
\includegraphics[scale=0.5]{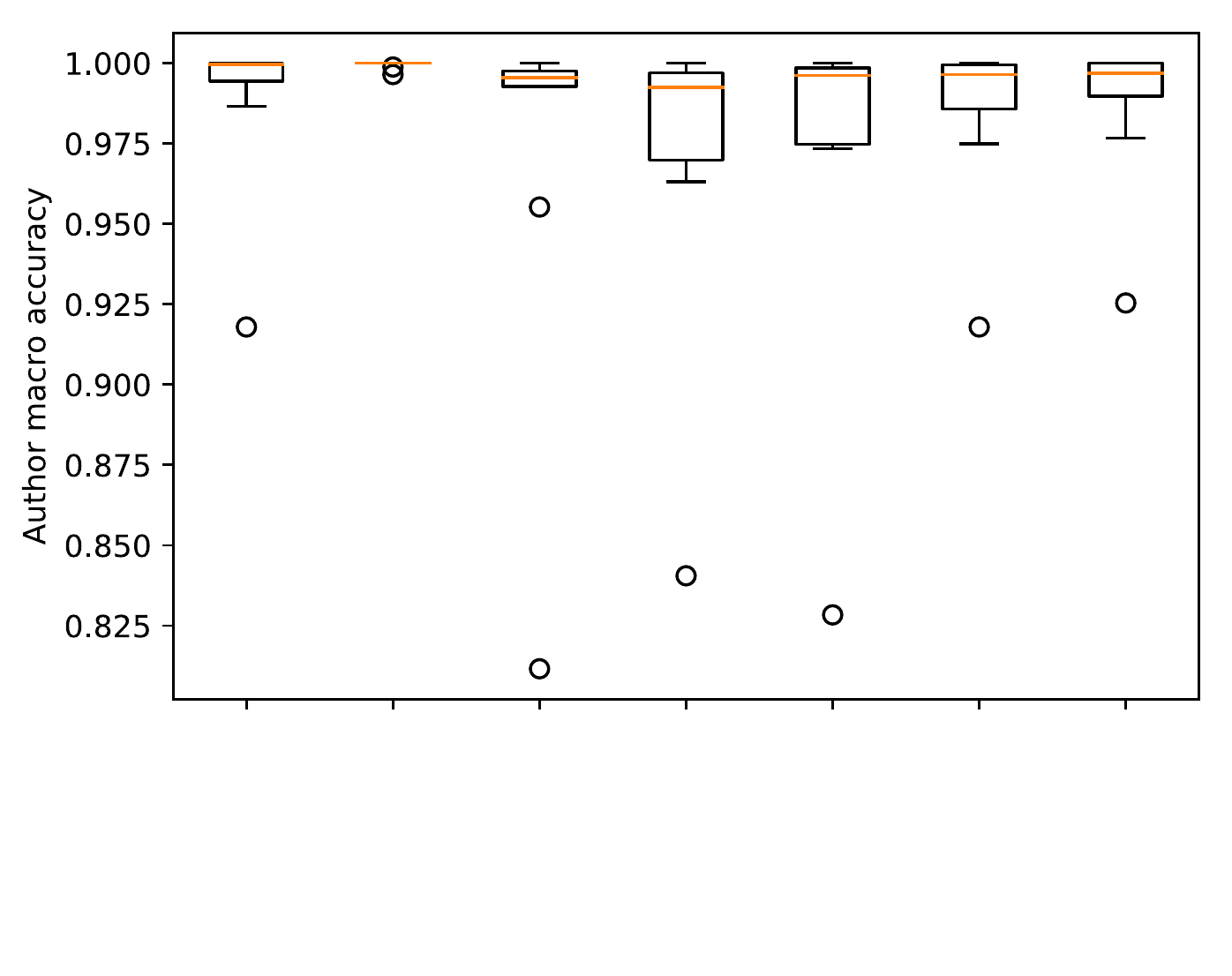}
\caption{normal test}
\label{fig:whisker-normal-theirs-sensitivity_to_correlation_changes}
\end{subfigure}
\begin{subfigure}{.5\textwidth}
\centering
\includegraphics[scale=0.5]{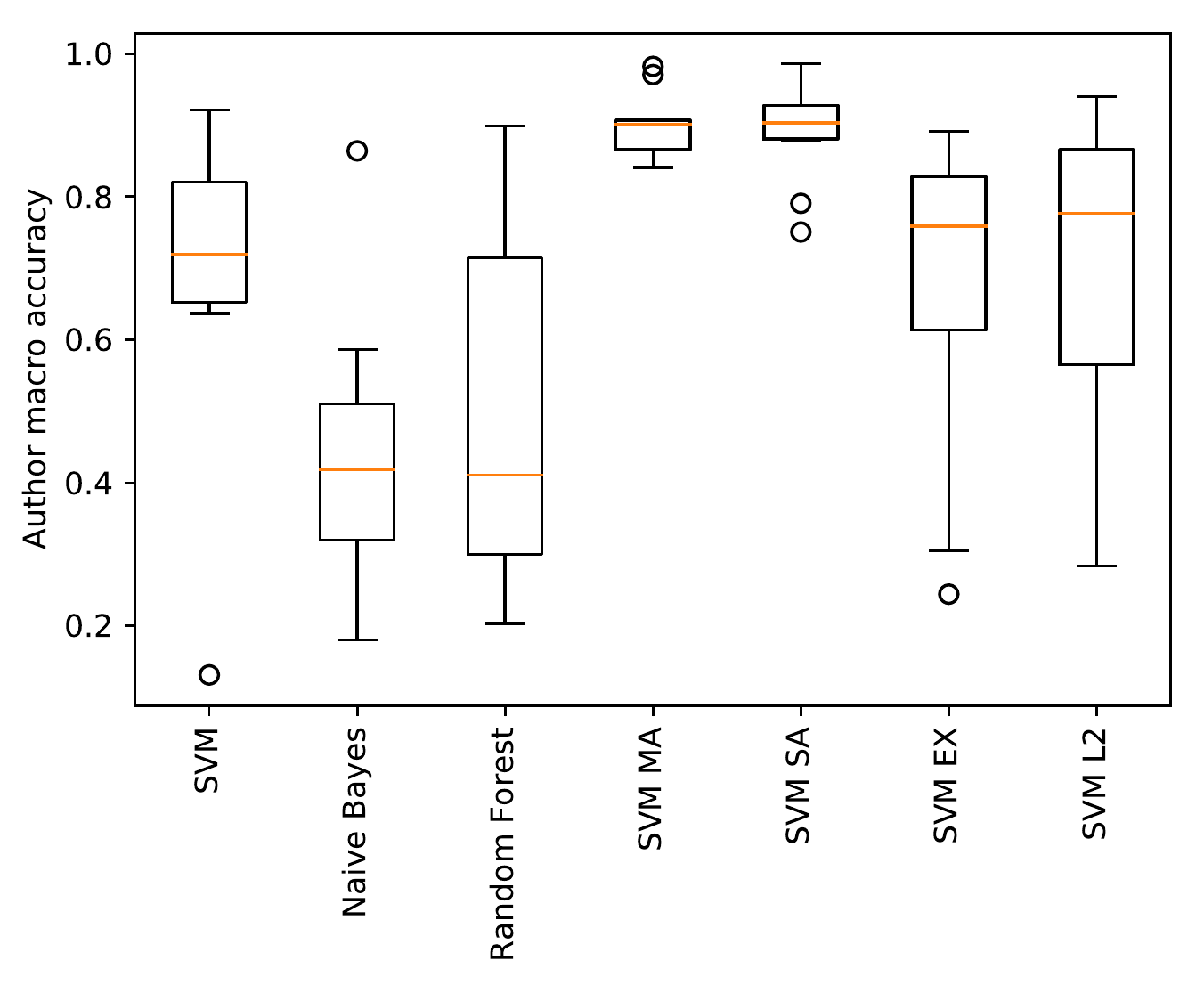}
\caption{swapped test}
\label{fig:whisker-swapped-theirs-sensitivity_to_correlation_changes}
\end{subfigure}
\caption{Author macro accuracy for (a) normal test set and (b) swapped test set as constructed in Section~\ref{sec:exp2}.}
\label{fig:whisker-theirs-sensitivity_to_correlation_changes}
\end{figure}

\paragraph{Zero-Knowledge Swapping}
Table~\ref{table-evaluation-results}b shows the accuracies of all except the Encoder models for zero-knowledge swapping alongside the accuracy delta between normal and swapped testing, and Figure~\ref{fig:whisker-theirs-sensitivity_to_correlation_changes} shows the respective performance distributions. The character trigram models which do not apply any measure to suppress domain style suffer severe drops of accuracy under swapping: The na{\"i}ve Bayes model, which under normal conditions achieves a perfect accuracy, drops~55.4 percentage points, falling even below random performance, so that reversing all of its decisions would yield a better performance. Similarly, the random forest-based and the SVM-based trigram models drop~46.8 and~31.5 percentage points, respectively.

\medskip
This leads us to the conclusion that the traditional way of building author style models is highly susceptible to learning domain style instead. Unless the domains are carefully controlled---which imposes severe practical limitations---these models are prone to pick up domain artifacts or be fooled by adversaries.

\medskip
Regarding the SVM-based models that apply heuristic rules to suppress domain style, their performance varies from similarly high drops in performance to much more sensible drops of~7.1 and~8.2 percentage points for the~MA and the~SA rules. Regarding the performance distributions in Figure~\ref{fig:whisker-theirs-sensitivity_to_correlation_changes}, in a normal test, all models perform quite consistently, whereas, in the swapped test, the performance distributions has a large variance with the exception of the SVM~MA and~SA.

\begin{figure}[t]
\begin{subfigure}{.5\textwidth}
\centering
\includegraphics[scale=0.5]{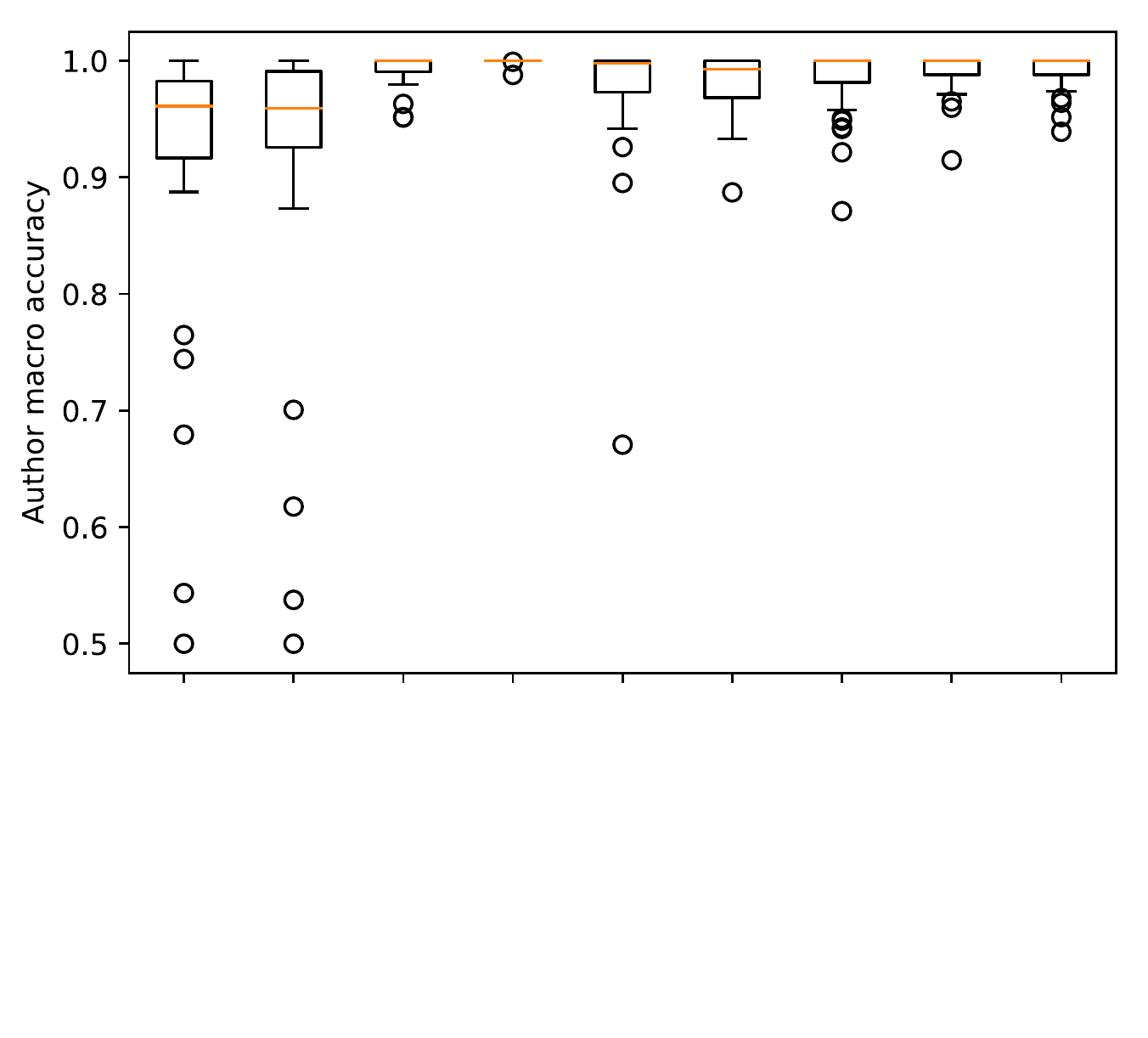}
\caption{normal test}
\label{fig:whisker-normal-relaxed_correlation_changes}
\end{subfigure}
\begin{subfigure}{.5\textwidth}
\centering
\vspace{1ex}
\includegraphics[scale=0.5]{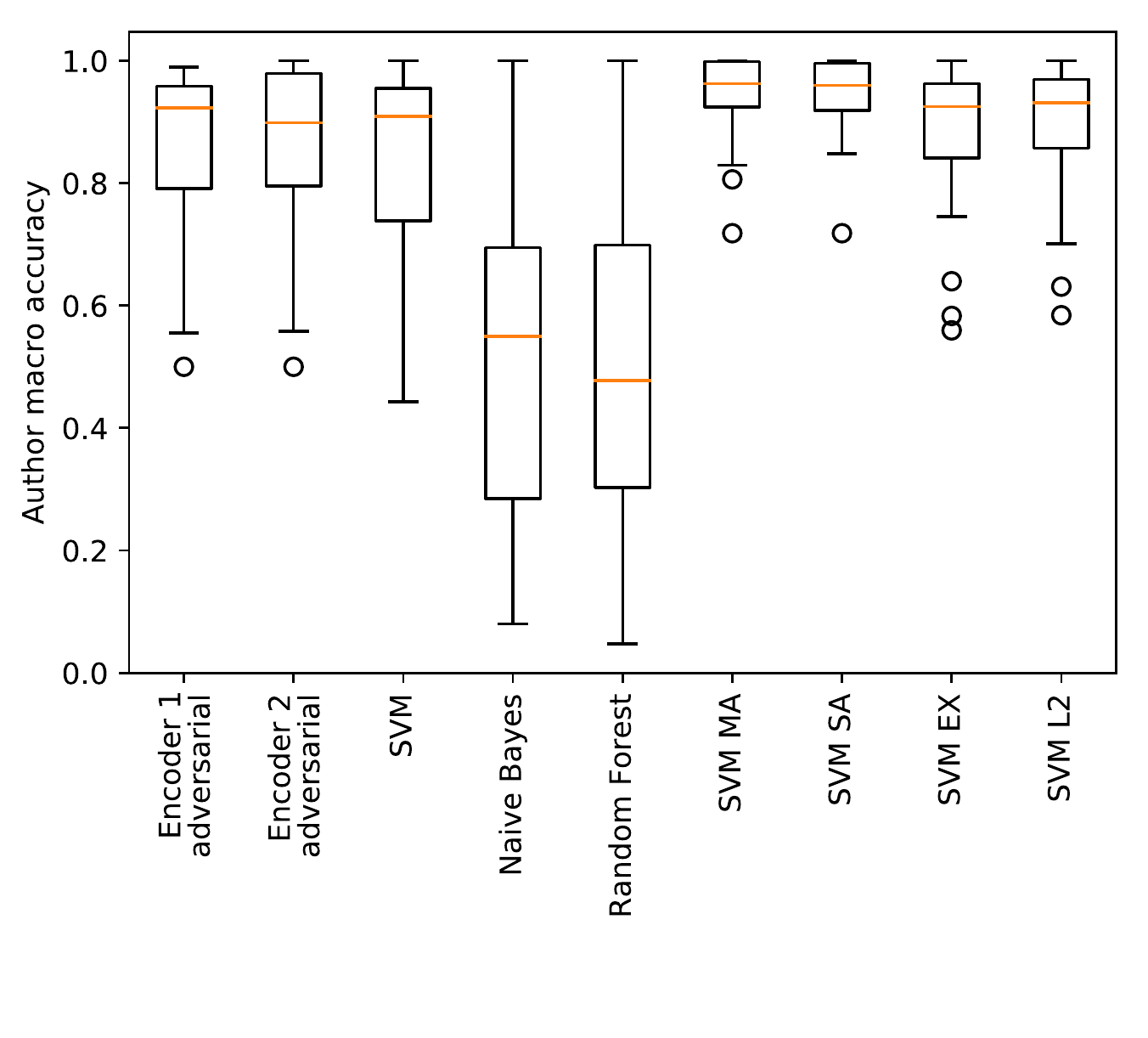}
\vspace{1ex}
\caption{swapped test}
\label{fig:whisker-swapped-relaxed_correlation_changes}
\end{subfigure}
\caption{Author macro accuracy for (a) normal test set and (b) swapped test set as constructed in Section~\ref{sec:exp2}.}
\label{fig:whisker-relaxed_correlation_changes}
\end{figure}

\paragraph{High-Imbalance Swapping}
Table~\ref{table-evaluation-results}c shows the accuracies of all models for high-imbalance swapping alongside the accuracy delta between normal and swapped testing, and Figure~\ref{fig:whisker-relaxed_correlation_changes} shows the respective performance distributions. 
The impact of domain swapping on the NB and RF models is comparable to that in the previous experiment. The SVM model, however, reduces its drop from~31.5 to only~15.4 percentage points. Consequently, also the SVM-based models that apply domain suppression drop much less than with zero-knowledge swapping. Regarding our Encoder models, their a priori performance under normal conditions more strongly deviates from the performance of the SVM-based models (around~90\% vs.\ 99\% accuracy). This is likely due to the effect of active domain suppression by adversarial training: The other models can exploit domain knowledge to achieve their high performance, whereas the Encoders cannot do so to the same extent. Moreover, the adversarial training component is perhaps penalized due to the high class imbalance. The drop in accuracy, between normal and swapped testing is the same as for the SVM-based models, which tells us that there is no relative disadvantage of adversarial training compared to heuristic rules: This is important, since it shows that domain style suppression can be learned instead of requiring handcrafted rules from experts.

\subsection{Cross-Fandom Authorship Attribution}
\label{exp:cross-fandom}

This experiment investigates the performance of all models within the following setup:

\input{table-experiment1a-schema}

\begin{figure}
\centering
\includegraphics[scale=0.5]{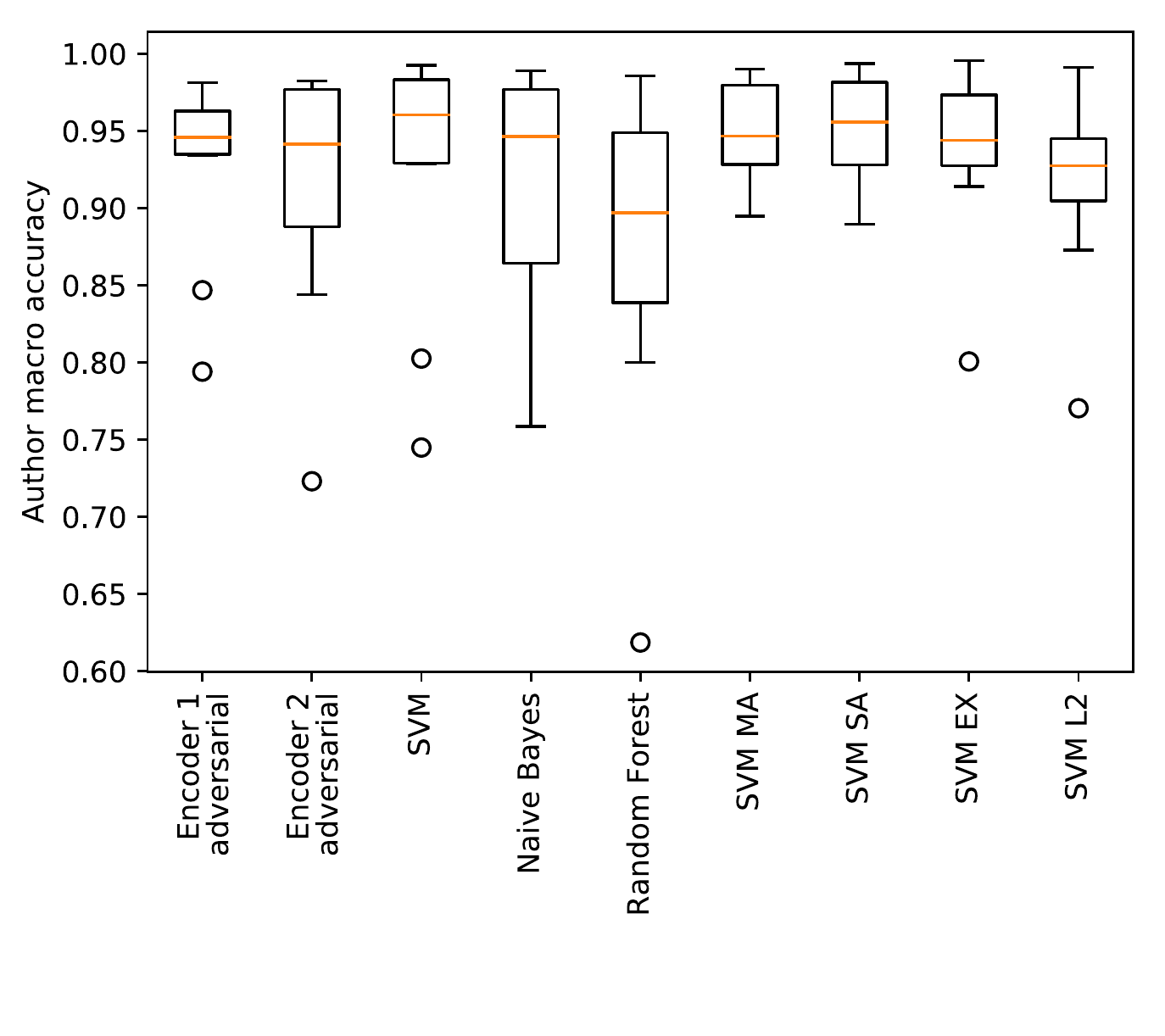}
\caption{Author macro accuracy for test set as constructed in Section~\ref{exp:cross-fandom}.}
\label{fig:whisker-cross_fandom_authorship_attribution}
\end{figure}

In this setup, two authors~A and~B have written in three fandoms~P, Q, and~R. The training set comprises an equal amount of text from both authors in the fandoms~P and~Q, while the test set is exclusively composed of equal amounts of text from both authors in fandom~R. This setup allows gives insights into the generalizability of the style encoder across fandoms, when the test fandom is unknown at the time of training. Table~\ref{table-evaluation-results}d shows the accuracies of all models for this setup, and Figure~\ref{fig:whisker-cross_fandom_authorship_attribution} shows the respective performance distributions. As can be seen, all models except for RF, tend to achieve a comparable accuracy of around~92\%, which shows that all models generalize across fandoms, and our style encoder in particular.

In a second generalization experiment, we investigate the performance of all models within the following setup:

\input{table-experiment1b-schema}

In this setup, two authors~C and~D have written in two fandoms~R and~S. However, the model is trained only equal amounts of text from author~C in both fandoms, while being tested on equal amounts of text from author~D in both fandoms. The main purpose of our proposed style encoder is to extract an author's writing style from given texts in a fandom-invariant manner: i.e., the fandom of a text must not be predictable, even if the network has not been trained on this specific fandom. In this experiment, we therefore take model trained for the experiment in Section~\ref{traditional-attribution} and extract the style of texts from new authors and fandoms. We then retrain the part of the network that uses the style vector to classify fandoms. The mean fandom accuracy for adversarially trained Encoder~1 is~51.6\% and that for Encoder~2 is~49.9\%, indicating the intended failure: The extracted style contains no usable fandom information.

The main function of our proposed network is that it extracts a writing style from given texts, which should be fandom-invariant, i.e., the fandom of a story must not be predictable, even if the network has not been trained on this specific fandom. For this purpose, we take the network trained in Section~\ref{traditional-attribution} and extract the style of texts from new authors and fandoms. We then retrain the part of the network that uses the style vector to classify fandoms, as shown in the following table:


%% file: table-evaluation-results.tex
\begin{table*}[tb]%
\centering%
\small%
\renewcommand{\tabcolsep}{7pt}%
\renewcommand{\arraystretch}{1}%
\begin{tabular}[t]{@{}l@{}}
\phantom{(a)} \\
\toprule
\bf Classifier \\
\\[1.1ex]
\midrule
SVM \\
NB \\
RF \\
\midrule
SVM MA\\
SVM SA\\
SVM EX\\ 
SVM L2\\ 
\midrule
Encoder~1\\ 
Encoder~2\\
\bottomrule
\end{tabular}%
\hfill%
\begin{tabular}[t]{@{}cc@{}}
\multicolumn{1}{@{}l}{(a)} \\
\toprule
\multicolumn{2}{@{}c@{}}{\bf Traditional Attribution } 
\\
\midrule
default & suppression  
\\
\midrule
96.8\%
& --
\\
%
95.3\%  
&--
\\ 
%
95.2\% 
& --
\\
\midrule
--
&  96.7\%
\\
%
--
&   97.3\%  
\\
%
--
& 96.4\%  
\\ 
%
--
&   96.9\%
\\ 
\midrule
95.5\% 
& 95.1\% 
\\ 
%
95.0\% 
&94.5\% 
\\ 
\bottomrule
\end{tabular}%
\hfill%
\begin{tabular}[t]{@{}ccc@{}}
\multicolumn{1}{@{}l}{(b)} \\
\toprule
\multicolumn{3}{@{}c@{}}{\bf Zero-Knowledge Swapping} \\
\midrule
normal & swapped  &  $\Delta$ 
\\
\midrule
98.9\% 
 & 67.4 \% 
 & $-$31.5\% \\ 
100\%
& 44.6 \% 
& $-$55.4\% \\ 
 97.4\%
 & 50.6\% 
 & $-$46.8\% \\ 
\midrule
 97.2\% 
 & 90.1\%
 & $-$7.1\%\\ 
97.4\% 
 & 89.2\%
 & $-$8.2\%\\
 98.6\% 
 & 67.3\% 
 & $-$31.3\%\\ 
98.8\%
 & 69.0\% 
 & $-$29.8\%\\ 
\midrule
\multicolumn{3}{@{}c@{}}{--}
\\
\multicolumn{3}{@{}c@{}}{--}
\\
\bottomrule
\end{tabular}%
\hfill%
\begin{tabular}[t]{@{}ccc@{}}
\multicolumn{1}{@{}l}{(c)} \\
\toprule
\multicolumn{3}{@{}c@{}}{\bf High-Imbalance Swapping} \\
\midrule
normal & swapped  &  $\Delta$ 
\\
\midrule
99.3\%
 &    83.9\% 
 & $-$15.4\% \\ 
100\% 
&     51.5\% 
& $-$48.5\% \\ 
 97.3\% 
 &     49.9\% 
 & $-$47.4\% \\ 
\midrule
 98.0\% 
 &     94.1\%
 & $-$3.9\%\\ 
98.1\%  
 &    94.5\% 
 & $-$3.6\%\\
 99.0\% 
 &     88.4\% 
 & $-$10.6\%\\ 
99.0\% 
 &    89.1\% 
 & $-$9.9\%\\ 
\midrule
90.8\% 
& 86.9\% 
& $-$3.9\% \\
91.0\% 
& 85.7\%
& $-$5.3\% \\ 
\bottomrule
\end{tabular}%
\hfill%
\begin{tabular}[t]{@{}c@{}}
\multicolumn{1}{@{}l}{(d)} \\
\toprule
\bf Section~\ref{exp:cross-fandom}  
\\
\\[1.1ex]
\midrule
 92.6\%  
\\
91.7\%  
\\ 
87.5\%
\\
\midrule
 94.8\%
\\
95.2\%  
\\
93.9\%   
\\ 
  91.7\%
\\ 
\midrule
92.8\% 
\\ 
 91.4\% 
\\ 
\bottomrule
\end{tabular}%
\caption{%
Mean macro accuracies on testing sets for classification of two authors who wrote in two fandoms.
(a)~Results of the experiment in Section~\ref{traditional-attribution}, where \textit{default} pertains to training the classifier to just predict the authors, and \textit{suppression} to training it in a way to reduce the domain style.
(b)~Results of the experiment in Section~\ref{sec:exp2}.
(c)~Results of the experiment in Section~\ref{sec:exp2}.
The results reported here for our network are based on adversarial training.
(d)~Results of the experiment in Section~\ref{exp:cross-fandom}
The results reported here for our network are based on adversarial training.
}%
\label{table-evaluation-results}%
\vspace{-7pt}
\end{table*}

%% file: table-experiment1-schema.tex
\begin{center}
\small
\setlength\tabcolsep{5pt}
\begin{tabular}{@{}r@{\qquad}cccc@{\qquad}cccc@{}}
\toprule
       & \multicolumn{4}{@{}c@{\qquad}}{training} & \multicolumn{4}{@{}c@{}}{test} \\ \midrule
author & A     & A    & B    & B    & A    & A    & B    & B   \\
fandom & P     & Q    & P    & Q    & P    & Q    & P    & Q   \\
\bottomrule
\end{tabular}
\end{center}

%% file: table-experiment2-schema.tex
\begin{center}
\small
\setlength\tabcolsep{5pt}
\begin{tabular}{@{}r@{\qquad}cc@{\qquad}cc@{\qquad}cc@{}}
\toprule
& \multicolumn{2}{@{}c@{\qquad}}{training}& \multicolumn{4}{@{}c@{}}{test} \\ \cmidrule(){4-7}
&&& \multicolumn{2}{@{}c@{\qquad}}{normal} & \multicolumn{2}{@{}c@{}}{swapped} \\ \midrule
author
& C & D            & C            & D           & C & D             \\
fandom
& R & S            & R            & S           & S & R             \\
\bottomrule
\end{tabular}
\end{center}

%% file: table-correlation-changes.tex
\begin{table}[t]
\centering
\small
\renewcommand{\tabcolsep}{3.0pt}
\begin{tabular}{@{}l @{\qquad} cc @{\qquad} cc @{\qquad} c@{}c@{}}
\addlinespace
\toprule
\addlinespace
& \multicolumn{2}{c@{\qquad}}{training} & \multicolumn{2}{c@{\qquad}}{normal test} & \multicolumn{2}{c}{swapped test} \\
\cmidrule(l{0pt}r{16pt}){2-3} \cmidrule(l{0pt}r{16pt}){4-5} \cmidrule(l{0pt}r{0pt}){6-7} \\[-1ex]
fandom   & R & S & R & S & R & S \\
\addlinespace
author A & 200 & 0 & max & 0 & 0 & max \\
author B & 0 & 200 & 0 & max & max & 0 \\
\bottomrule
\end{tabular}
\caption{Zero-knowledge domain swapping: During training, a model has no access to the one relation between authors and fandoms, whereas during swapped testing, the situation is reversed.}
\label{table-correlation-changes}
\end{table}

%% file: table-correlation-changes-relaxed.tex
\begin{table}[t]
\centering
\small
\renewcommand{\tabcolsep}{3.0pt}
\begin{tabular}{@{}l @{\qquad} cc @{\qquad} cc @{\qquad} c@{}c@{}}
\addlinespace
\toprule
\addlinespace
& \multicolumn{2}{c@{\qquad}}{training} & \multicolumn{2}{c@{\qquad}}{normal test} & \multicolumn{2}{c}{swapped test} \\
\cmidrule(l{0pt}r{16pt}){2-3} \cmidrule(l{0pt}r{16pt}){4-5} \cmidrule(l{0pt}r{0pt}){6-7} \\[-1ex]
fandom   & T & U & T & U & T & U \\
\addlinespace
author E & 600 & 10 & max & 0 & 0 & max \\
author F & 10 & 600 & 0 & max & max & 0 \\
\bottomrule
\end{tabular}
\caption{High-imbalance domain swapping: During training, a model has highly imbalanced access to both relations between authors and fandoms, whereas during swapped testing, the situation is reversed.}
\label{table-correlation-changes-relaxed}
\end{table}

%% file: table-experiment1a-schema.tex
\begin{center}
\small
\setlength\tabcolsep{5pt}
\begin{tabular}{@{}r@{\qquad}cccc@{\qquad}cc@{}}
\toprule
        & \multicolumn{4}{@{}c@{\qquad}}{training} & \multicolumn{2}{@{}c@{}}{test} \\ \midrule
author & A     & A    & B    & B    & A    & B      \\
fandom & P     & Q    & P    & Q    & R    & R      \\
\bottomrule
\end{tabular}
\end{center}

%% file: table-experiment1b-schema.tex
\begin{center}
\small
\setlength\tabcolsep{5pt}
\begin{tabular}{@{}r@{\qquad}cc@{\qquad}cc@{}}
\toprule
             & \multicolumn{2}{@{}c@{\qquad}}{\kern-0.5em retraining} & \multicolumn{2}{@{}c@{}}{test} \\\hline
author          
& C            & C           & D & D              \\
fandom 
& R            & S           & R & S              \\
\bottomrule
\end{tabular}
\end{center}

%% file: domain-style-sum.tex
\section{Conclusion and Future Work}

Representing writing style poses many theoretical and practical problems: Not only is there no clear definition of what, precisely, constitutes an author's writing style, and what separates it from other kinds of writing style due to domains like genre, register, and topic; as we reveal in this paper: the traditional way of modeling writing style is highly susceptible to representing domain style instead of author style. This revelation is due to a new kind of experimental setup that is afforded by a new, large-scale corpus that we build as part of this work. The corpus comprises large amounts of long texts written by many different authors in many different domains, allowing for the first time to study the relation of author style and domain style for at least one domain of interest.

We demonstrate that basic models employing character trigrams as features are reduced to near-random performance under our experimental setup. We further show that two approaches to reduce domain style yield promising results: on the one hand, a heuristic approach of handcrafted rules to select features, and on the other, a new domain-adversarial learning approach that learns to extract writing style while suppressing domain style. Both appear to be working equally well. However, our adversarial learning approach is presumably more adaptable to new situations.

On a more critical note, the amount of data required to train our model to high performance is considerable, and it stands to reason that such amounts of data cannot be easily compiled in practical situations, and even less so when more than one domain has to be suppressed. In this regard, the good performance of the handcrafted rules at least open the door to a less resource-intensive approach. Nevertheless, the amount of training data required to train a neural network may border on impracticality, unless we can create pre-trained style models that need only be fine-tuned, e.g., like BERT. Future work in this direction is necessary, and perhaps the directions we have shown as well as the data we have compiled will help to go into this direction.